\documentclass[sigconf]{acmart}

\AtBeginDocument{%
  \providecommand\BibTeX{{%
    \normalfont B\kern-0.5em{\scshape i\kern-0.25em b}\kern-0.8em\TeX}}}

\copyrightyear{2023}
\acmYear{2023}
\setcopyright{acmlicensed}\acmConference[KDD '23]{Proceedings of the 29th ACM SIGKDD Conference on Knowledge Discovery and Data Mining}{August 6--10, 2023}{Long Beach, CA, USA}
\acmBooktitle{Proceedings of the 29th ACM SIGKDD Conference on Knowledge Discovery and Data Mining (KDD '23), August 6--10, 2023, Long Beach, CA, USA}
\acmPrice{15.00}
\acmDOI{10.1145/3580305.3599439} \acmISBN{979-8-4007-0103-0/23/08}

\usepackage{multirow}
\usepackage[normalem]{ulem}
\usepackage{MnSymbol}
\usepackage{float}
\useunder{\uline}{\ul}{}
\usepackage{appendix}
\usepackage{hyperref}
\usepackage{enumitem}
\usepackage{stfloats}
\usepackage{xcolor}

\begin{document}

\title{Multi-Grained Multimodal Interaction Network for Entity Linking}

\author{Pengfei Luo}
\orcid{0000-0003-0889-7660}
\affiliation{%
  \institution{School of Computer Science and
Technology, University of Science and
Technology of China \& State Key
Laboratory of Cognitive Intelligence}
  \city{Hefei}
  \state{Anhui}
  \country{China}
}
\email{pfluo@mail.ustc.edu.cn}

\author{Tong Xu}
\orcid{0000-0003-4246-5386}
\affiliation{%
  \institution{School of Computer Science and
Technology, University of Science and
Technology of China \& State Key
Laboratory of Cognitive Intelligence}
  \city{Hefei}
  \state{Anhui}
  \country{China}
}
\email{tongxu@ustc.edu.cn}
\authornote{Corresponding Author.}

\author{Shiwei Wu}
\orcid{0000-0002-3206-6827}
\affiliation{%
  \institution{School of Data Science, University of Science and
Technology of China \& State Key
Laboratory of Cognitive Intelligence}
  \city{Hefei}
  \state{Anhui}
  \country{China}
}
\email{dwustc@mail.ustc.edu.cn}

\author{Chen Zhu}
\orcid{0000-0003-4817-482X}
\affiliation{%
  \institution{Career Science Lab, BOSS Zhipin \& School of Management, University of Science and Technology of China}
  \city{Beijing}
  \country{China}
}
\email{zc3930155@gmail.com}

\author{Linli Xu}
\orcid{0000-0003-0227-3793}
\affiliation{%
  \institution{School of Computer Science and
Technology, University of Science and
Technology of China \& State Key
Laboratory of Cognitive Intelligence}
  \city{Hefei}
  \state{Anhui}
  \country{China}
}
\email{linlixu@ustc.edu.cn}

\author{Enhong Chen}
\orcid{0000-0002-4835-4102}
\affiliation{%
  \institution{School of Computer Science and
Technology, University of Science and
Technology of China \& State Key
Laboratory of Cognitive Intelligence}
  \city{Hefei}
  \state{Anhui}
  \country{China}
}
\email{cheneh@ustc.edu.cn}
\authornotemark[1]

\renewcommand{\shortauthors}{Pengfei Luo et al.}

\begin{abstract}
  Multimodal entity linking (MEL) task, which aims at resolving ambiguous mentions to a multimodal knowledge graph, has attracted wide attention in recent years. Though large efforts have been made to explore the complementary effect among multiple modalities, however, they may fail to fully absorb the comprehensive expression of abbreviated textual context and implicit visual indication. Even worse, the inevitable noisy data may cause inconsistency of different modalities during the learning process, which severely degenerates the performance. To address the above issues, in this paper, we propose a novel Multi-GraIned Multimodal InteraCtion Network \textbf{(MIMIC)} framework for solving the MEL task. Specifically, the unified inputs of mentions and entities are first encoded by textual/visual encoders separately, to extract global descriptive features and local detailed features. Then, to derive the similarity matching score for each mention-entity pair, we device three interaction units to comprehensively explore the intra-modal interaction and inter-modal fusion among features of entities and mentions. In particular, three modules, namely the Text-based Global-Local interaction Unit (TGLU), Vision-based DuaL interaction Unit (VDLU) and Cross-Modal Fusion-based interaction Unit (CMFU) are designed to capture and integrate the fine-grained representation lying in abbreviated text and implicit visual cues. Afterwards, we introduce a unit-consistency objective function via contrastive learning to avoid inconsistency and model degradation. Experimental results on three public benchmark datasets demonstrate that our solution outperforms various state-of-the-art baselines, and ablation studies verify the effectiveness of designed modules \footnote{Our code is available at \url{https://github.com/pengfei-luo/MIMIC}}.

\end{abstract}




\maketitle

\section{Introduction}

\begin{figure}
    \centering
    \includegraphics[width=0.46\textwidth]{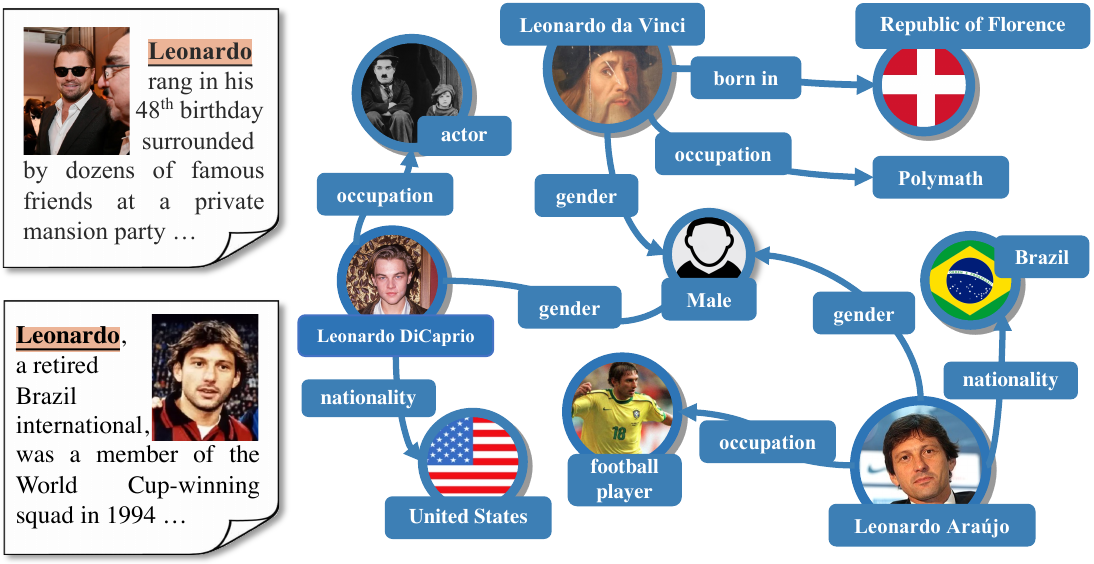}
    \caption{Examples of multimodal entity linking. Left: two multimodal mentions. Right: multimodal knowledge graph.}
    \label{fig:introduction}
\end{figure}

Entity linking (EL), also known as entity disambiguation, plays a fundamental but imperative role to connect a wide and diverse variety of web content to referent entities of a knowledge graph (KG), which supports numerous downstream applications such as search engines~\cite{DBLP:conf/sigir/GerritseHV22, DBLP:conf/cidr/ChengC07}, question answering~\cite{DBLP:conf/emnlp/LongprePCRD021, DBLP:conf/acl/XiongYCGW19}, dialog systems~\cite{DBLP:conf/cikm/AhmadvandSCA19, DBLP:conf/mm/LiaoM0HC18} and so on. 
Over the past years, large efforts have been dedicated to text-based entity linking. However, in the surge of multimodal information, images along with text have become the most widely-seen medium to publishing and understanding web information, which also brings challenges to the comprehension of complex multimodal content. Thereby, multimodal entity linking (MEL), resolving the visual and textual mentions into their corresponding entities of a multimodal knowledge graph (MMKG), is desperately desired. For instance, as shown in Figure~\ref{fig:introduction}, the short sentence contains an affiliated image to complement the textual context of mention. In this case, it is challenging for text-based EL methods to determine which entity is related to the entity \textit{Leonardo} in Figure~\ref{fig:introduction}. Differently, visual information, e.g., the character portraits, brings valuable content and alleviates ambiguity of textual modality. Thus, it is intuitive to integrate visual information with textual contexts when linking the multimodal mentions to heterogeneous MMKG entities.

Along this line, prior arts attempted to solve the MEL task via exploring complementary effects of different modalities by leveraging concatenation operation~\cite{DBLP:conf/ecir/AdjaliBFBG20},  additive attention~\cite{DBLP:conf/acl/CarvalhoMN18}, and cross-attention mechanism~\cite{DBLP:conf/sigir/WangWC22} on public benchmark datasets such as TwitterMEL~\cite{DBLP:conf/ecir/AdjaliBFBG20}, WikiMEL~\cite{DBLP:conf/sigir/WangWC22}. Although these studies for MEL have shown promising progress compared with text-based EL methods, MEL is still not a trivial task due to the following reasons: 
\begin{enumerate}[leftmargin=*]
    \item \textbf{Short and abbreviated textual context}. The sentence of mention contexts contains limited information due to text length or the known topic, which is commonly seen in social media platforms. Therefore, it is necessary to capture the fine-grained clues lay in the textual context.
    \item \textbf{Implicit visual indication}. Due to the ``semantic gap” between low-level visual information and high-level semantic cues, it might be difficult to capture the implicit indications that correspond to the category or description of entities. For example, the portrait could imply occupation and gender of one person, which may not be extracted via simple detection or matching tools. In this case, it is necessary to design one specific module to capture the implicit multimodal cues from explicit visual features.
    \item \textbf{Modality Consistency}. Recent studies~\cite{DBLP:conf/mm/ChenDLZH19, DBLP:conf/kdd/ChenL00WYC22} have revealed that joint learning of multiple modalities may cause contradiction or degeneration when optimization due to the inevitable noisy data, or excessive influence of a specific modality. Therefore, it is necessary to model consistency and enhance the cooperative effect among modalities.
\end{enumerate}

To deal with these issues, in this paper, we propose a novel \textbf{M}ulti-Gra\textbf{I}ned \textbf{M}ultimodal \textbf{I}ntera\textbf{C}tion network \textbf{(MIMIC)} for MEL task, which consists of two layers, namely an input and feature encoding layer, as well as a multi-grained multimodal interaction layer. Specifically, in the input and feature encoding layer, we design a unified input format for both multimodal mention and MMKG entities. Then, the encoder extracts both local and global features of textual and visual inputs for obtaining global descriptive semantics, while reserving fine-grained details in words or image patches. Also, in the multi-grained multimodal interaction layer, we devise three parallel interaction units to fully explore multimodal schemata. First, to capture the clues that lie in the abbreviated text, we propose a Text-based Global-Local interaction Unit (TGLU), which not only considers lexical coherence from a global view but also mines fine-grained semantics by utilizing attention mechanism. Afterwards, to address the challenge of visual indication, we design a Vision-based DuaL interaction Unit (VDLU) and a Cross-Modal Fusion-based interaction Unit (CMFU), for explicit and implicit indications, respectively. In detail, the tailored VDLU introduces a dual-gated mechanism to amplify the explicit visual evidence within features as well as enhance robustness against noisy images from the Internet. Meanwhile, different from utilizing concatenation or attention, the CMFU module first projects extracted global textual features and local visual features into a vector space, and then fuse them with a gated operation, which could effectively mine the implicit semantic relevance of multiple modalities to complement each other. Moreover, to attain the consistency of different modalities and units, we introduce a unit-consistent loss function based on contrastive training to improve intra-modal and inter-modal learning for multiple interaction units. To the best of our knowledge, technical contributions of this paper can be summarized as follows:
\begin{itemize}[leftmargin=*]
    \item We propose a multi-grained multimodal interaction network for solving multimodal entity linking task, which could universally extract features for both multimodal mentions and entities. And the proposed network could be easily extended by adding new interaction units.
    \item We devise three interaction units to sufficiently explore and extract diverse multimodal interactions and patterns for entity linking. Moreover, we introduce the unit-consistent loss function to enhance the intra-modal and inter-modal representation learning.
    \item We perform extensive experiments on three public multimodal entity linking datasets. Experimental results illustrate that our methods outperform various competitive baselines. The ablation study also validates the effectiveness of each designed module.
\end{itemize}

\section{Related Work}
The related methods can be categorized into text-based entity linking and multimodal entity linking based on the modalities they use. We elaborate on them one after the other.

\subsection{Text-based Entity Linking}
This line of research links mentions to a known knowledge graph via utilizing textual information of context and entities. According to  the granularity of different methods, we roughly divide the existing studies into two groups: local-level methods and global-level methods. The former approaches primarily perform entity linking by mapping mention along with its surrounding words or sentence for similarity calculation. Early research leveraged word2vec and convolutional neural networks (CNN) to capture the correlation between mention context and entity information~\cite{DBLP:conf/conll/YamadaS0T16, DBLP:conf/naacl/Francis-LandauD16, DBLP:conf/naacl/TsaiR16, DBLP:conf/acl/CaoHJCL17}.
Thereafter, Eshel et al.~\cite{DBLP:conf/conll/EshelCRMYL17} integrated entity embedding into the recurrent neural network (RNN) with attention mechanism in order to exploit the sequential nature of the noisy and short context. To mine the diverse entity-side external information, Gupta et al.~\cite{DBLP:conf/emnlp/GuptaSR17} further explored the fusion among entity description and fine-grained entity type for robust and meaningful  representations.
Motivated by the popularity of Transformer~\cite{DBLP:conf/nips/VaswaniSPUJGKP17}, Peters et al.~\cite{DBLP:conf/emnlp/PetersNLSJSS19} designed a projection layer over mention spans, and recontextualized these spans with cross-attention to link entity as well as integrate knowledge into BERT~\cite{DBLP:conf/naacl/DevlinCLT19}. In addition, Wu et al.~\cite{DBLP:conf/emnlp/WuPJRZ20} developed a two-stage linking algorithm towards the zero-shot scenario. They employed BERT to encode entities and mention context separately and then utilized a Transformer layer for detailed context-candidate scoring. By contrast,  De Cao et al.~\cite{DBLP:conf/iclr/CaoI0P21} modeled entity linking in an auto-regressive manner by using BART~\cite{DBLP:conf/acl/LewisLGGMLSZ20} architecture to generate the unique names of different entities.

The latter stream mainly tries to disambiguate several entity occurrences from a document-global view and takes into consideration semantic consistency as well as entity coherence, which also leads to high computation complexity. As one of the representative studies, Le and Titov~\cite{DBLP:conf/acl/TitovL18a} proposed to encode relations among different mentions as latent variables, and induced them with a multi-relational neural model. Based on the assumption that previously identified entities bring cues for the subsequent linking, Fang et al.~\cite{DBLP:conf/www/FangC0ZZL19} treated entity linking as the sequential decision problem and resolved it with reinforcement learning. At the same period,  Yang et al.~\cite{DBLP:conf/emnlp/YangGLTZWCHR19} extended this paradigm by accumulating attributes from the previously linked entities to enhance the decoding procedure. Another thread of this line constructs all mentions as nodes of a graph and uses the similarity of different nodes as edges. Thereinto, Cao et al.~\cite{DBLP:conf/coling/0002HLL18} employed graph convolution network (GCN) to integrate features and global coherence. Besides, Wu et al.~\cite{DBLP:conf/www/WuZMGSH20} proposed a dynamic GCN architecture to alleviate the insufficiency of structural information. 

Although text-based methods have achieved significant progress, they usually ignore the critical and abundant visual information of vivid images, which results in the failure to integrate visual cues.

\subsection{Multimodal Entity Linking} 
Since social media and news posts are in the form of texts and images, combining both textual and visual information for entity linking is crucial and practical. As one of the pioneering research, Moon et al.~\cite{,DBLP:conf/acl/CarvalhoMN18} introduced images to assist entity linking due to the polysemous and incomplete mentions from social media posts. Beyond that, Adjali et al.~\cite{DBLP:conf/ecir/AdjaliBFBG20} utilized unigram and bigram embeddings as textual features and pretrained Inception~\cite{DBLP:conf/cvpr/SzegedyVISW16} to extract visual features. After the extraction, a concatenation operation was applied to fuse the features and the model was optimized with the triple loss. They also constructed a MEL dataset of social media posts from Twitter. Wang et al.~\cite{DBLP:conf/sigir/WangWC22} further explored inter-modal correlations via a text and vision cross-attention, where a gated hierarchical structure is incorporated. To remove the negative effect caused by noisy and irrelevant images, Zhang et al.~\cite{DBLP:conf/dasfaa/ZhangLY21} considered the correlation between the category information of images and the semantic information of text mentions, in which the images were filtered by a predeﬁned threshold. Gan et al.~\cite{DBLP:conf/mm/GanLWWHH21} constructed a dataset that contains long movie reviews with various related entities and images. A recent research~\cite{DBLP:journals/dint/ZhengWWQ22} incorporated scene graphs of images to obtain object-level encoding towards detailed semantics of visual cues.  

Although these research studies have shown that visual information is beneficial to the performance of entity linking to some extent,  the utilization of visual information in conjunction with textual context remains largely underdeveloped.

\section{Methodology}
In this section, we first formulate the task of multimodal entity linking, and then go through the details of the proposed framework.

\subsection{Problem Formulation}
First, we define related mathematical notations as follows. Typically, a multimodal knowledge base is constructed by a set of entities $\mathcal{E}= \left \{\mathbf{E}_i \right \}_{i=1}^{N}$, and each entity is denoted as $\mathbf{E}_i=(\mathbf{e}_{n_i}, \mathbf{e}_{v_i}, \mathbf{e}_{d_i}, \mathbf{e}_{a_i})$, where the elements of $\mathbf{E}_i$ represent entity name, entity images, entity description, and entity attributes, respectively. Since our research concentrates on local-level entity linking, the textual inputs are in the format of sentences instead of documents. Here, a mention and its context are denoted as $\mathbf{M}_j=(\mathbf{m}_{w_j}, \mathbf{m}_{s_j}, \mathbf{m}_{v_j})$, where $\mathbf{m}_{w_j}, \mathbf{m}_{s_j}$ and $\mathbf{m}_{v_j}$ indicate the words of mention, the sentence in which the mention is located, and the corresponding image, respectively. The related entity of the mention $\mathbf{M}_j$ in the knowledge base is $\mathbf{E}_i$.

Along this line, given a mention $\mathbf{M}_j$, the task of multimodal entity linking targets to retrieve the ground truth entity $\mathbf{E}_i$ from the entity set $\mathcal{E}$ of knowledge base. This task can be obtained by maximizing the log-likelihood over the training set $\mathcal{D}$ while optimizing the model parameters $\theta$, i.e., 
\begin{equation}
    \theta^* = \mathop{\max}\limits_{\theta} \sum_{(\mathbf{M}_j, \mathbf{E}_i) \in \mathcal{D}} \log p_{\theta} \left ( \mathbf{E}_i | \mathbf{M}_j, \mathcal{E} \right ),
\end{equation}
where $\theta^*$ indicates the final parameters. Afterwards, we resolve $p_{\theta}(\mathbf{E}_i | \mathbf{M}_j, \mathcal{E})$ via calculating the similarity between the mention and each entity of the given knowledge base.

\subsection{Input and Encoding Layer}
In this layer, we design a unified input format, which allows mentions and entities to share the same visual/textual encoder. We introduce the input format and encoding process in the following subsections.
\begin{figure}
    \centering
    \includegraphics[width=0.475\textwidth]{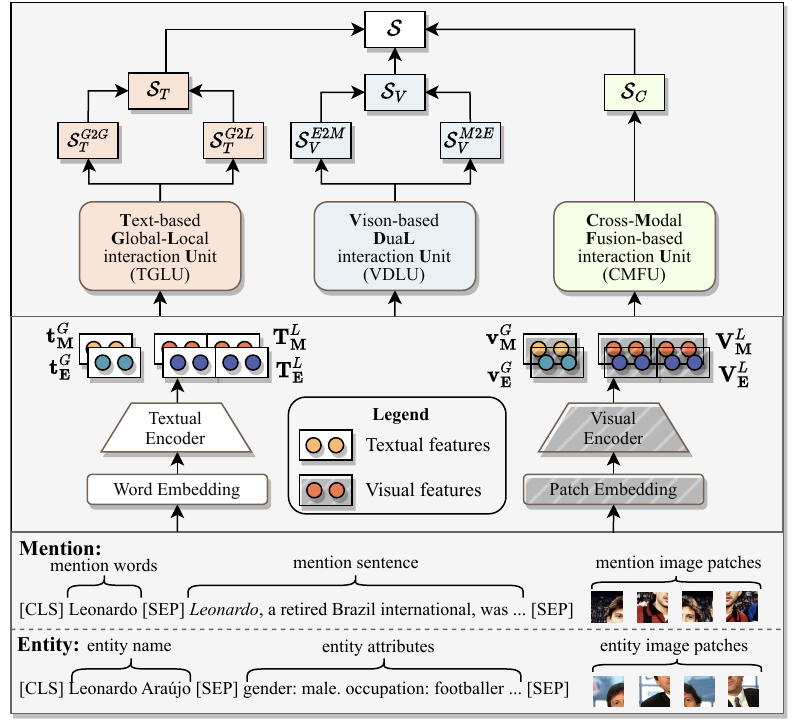}
    \caption{An overview of MIMIC. The bottom part is the input layer. The middle part is the encoding layer. The upper part is the multi-grained multimodal interaction layer.}
    \label{fig:overview}
\end{figure}
\subsubsection{\textbf{Visual Feature Encoding}}
To capture the expressive features of images, we employ the pre-trained Vision Transformer (ViT)~\cite{DBLP:conf/iclr/DosovitskiyB0WZ21} as the visual encoder backbone. Given the image $\mathbf{e}_{v_i}$ of an entity $\mathbf{E}_i$, we first rescale each image into $C \times H \times W$ pixels and reshape it into $n = H \times W /P^2$ flattened 2D patches, where $C$ is the number of channels, $H \times W$ is the image resolution and $P$ represents the patch size. After that, the patches go through the projection layer and multi-layer transformer of the standard ViT. We add a fully connected layer to convert the dimension of output hidden status into $d_v$. Thus the hidden status of entity image are denoted as $\mathbf{V}_{\mathbf{E}_i} = \left [ \mathbf{v}_{\text{[CLS]}}^0; \mathbf{v}_{\textbf{E}_i}^1; \dots; \mathbf{v}_{\mathbf{E}_{i}}^n \right ] \in \mathbb{R}^{(n + 1) \times d_v}$. We take the corresponding hidden state of the special token \text{[CLS]} as global feature $\mathbf{v}_{\mathbf{E}_i}^G \in \mathbb{R}^{d_v}$ and the whole hidden states as local features  $\mathbf{V}_{\mathbf{E}_i}^L \in \mathbb{R}^{(n+1) \times d_v}$. Similarly, for the image of mention $\mathbf{\mathbf{M}_j}$, we obtain $\mathbf{v}_{\mathbf{M}_j}^G$ as global visual feature and $\mathbf{V}_{\mathbf{M}_j}^L$ as local visual features.

\subsubsection{\textbf{Textual Feature Encoding}}
To extract meaningful word embeddings, we utilize a pre-trained BERT~\cite{DBLP:conf/naacl/DevlinCLT19} as the textual encoder. We construct the input of an entity by concatenating the entity name with its attributes, i.e., 
\begin{equation}
    \mathbf{I}_{\mathbf{E}_i} = \text{[CLS]} \mathbf{e}_{n_i} \text{[SEP]} \mathbf{e}_{a_i} \text{[SEP]},
\end{equation}
where $e_{a_i}$ is a set of entity attributes collected from the knowledge base including entity type, occupation, gender, and so on. Different attributes are separated by a period. Then we feed the tokenized sequence $\mathbf{I}_{\mathbf{E}_i}$ into BERT and the hidden states are denoted as $\mathbf{T_{\mathbf{E}_i}} = \left [ \mathbf{t}_{\text{[CLS]}}^0; \mathbf{t}_{\mathbf{E}_i}^1; \dots; \mathbf{t}_{\mathbf{E}_i}^{l_e} \right ] \in \mathbb{R}^{(l_e + 1) \times d_t}$, where $d_T$ is the dimension of textual output features, and $l_e$ is the length. We also regard the hidden state of \text{[CLS]} as global textual feature $\mathbf{t}_{\mathbf{E}_i}^G$ and the entire hidden states $\textbf{T}_{\textbf{E}_i}$ as local textual features $\mathbf{T}_{\mathbf{E}_i}^L$.

As for the mention $\textbf{M}_j$, we use the concatenation of the words of mention and the sentence where the mention is located to compose the input sequence. This can be illustrated as,
\begin{equation}
    \mathbf{I}_{\mathbf{M}_j} = \text{[CLS]} \mathbf{m}_{w_j} \text{[SEP]} \mathbf{m}_{s_j} \text{[SEP]}.
\end{equation}
Similarly, following the procedure that we process entity, we also obtain $\mathbf{t}_{\mathbf{M}_j}^G$ and $\mathbf{T}_{\mathbf{M}_j}^L$ as local textual features and global textual features of the mention $\mathbf{M}_j$ respectively. Notably, in the following subsection, we drop the subscript $i$ of entity and $j$ of mention for mathematical conciseness.

\subsection{Multi-Grained Multimodal Interaction Layer}
\begin{figure*}[ht]
    \centering
    \includegraphics[width=0.99\textwidth]{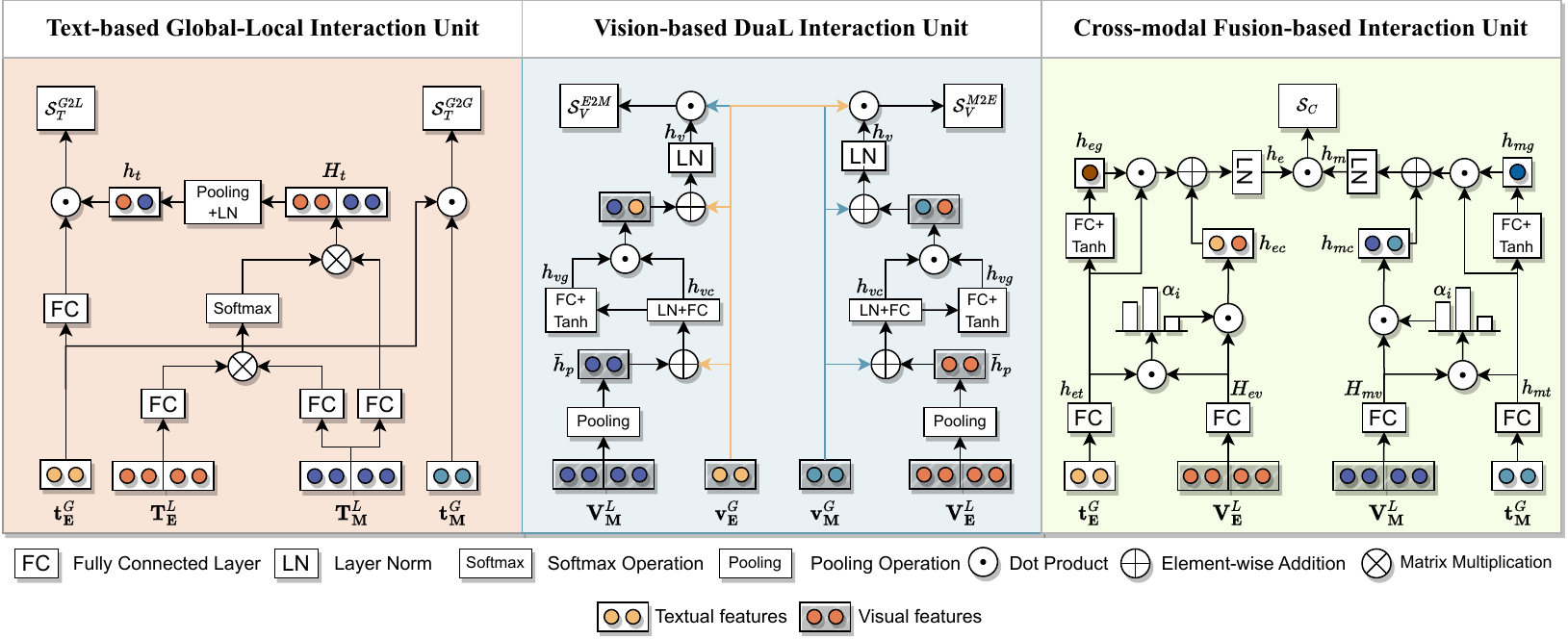}
    \caption{The designed multi-grained multimodal interaction layer, which contains three interaction units.}
    \label{fig:model}
\end{figure*}
To derive similarity matching scores for each mention-entity pair, we devise three interaction units by fully exploring the intra-modal and inter-modal clues in different granularities. As illustrated in Figure~\ref{fig:model}, the interaction layer consists of three parallel units: (1) \textbf{Text-based Global-Local interaction Unit} (TGLU) is dedicated to capturing lexical information among abbreviated text in both whole and partial views; (2) \textbf{Vision-based DuaL interaction Unit} (VDLU) concentrates on revealing the explicit visual correlation between mention images and entity images; (3) \textbf{Cross-Modal Fusion-based interaction Unit} (CMFU) focuses on capturing fine-grained implicit semantics to supplement the interaction of different modalities. Each unit takes features from an entity and a mention as inputs and then calculates a score as:
\begin{gather}
    \mathcal{S}_T = \mathcal{U}_T \left( \mathbf{M}, \mathbf{E} \right) = ( \mathcal{S}_T^{G2G} + \mathcal{S}_T^{G2L} ) / 2, \label{eq:ST} \\
    \mathcal{S}_V =  \mathcal{U}_V \left( \mathbf{M}, \mathbf{E} \right) = ( \mathcal{S}_V^{E2M} + \mathcal{S}_V^{M2E} ) / 2, \label{eq:SV} \\
    \mathcal{S}_C = \mathcal{U}_C \left( \mathbf{M}, \mathbf{E}  \right), \label{eq:SC} \\
    \mathcal{S} = \mathcal{U} \left( \mathbf{M}, \mathbf{E}  \right) = ( \mathcal{S}_V + \mathcal{S}_T + \mathcal{S}_C ) / 3, \label{eq:S}
\end{gather}
where $\mathcal{S}_T$, $\mathcal{S}_V$, and $\mathcal{S}_C$ are the scores calculated by TGLU, VDLU, and CMFU respectively. The final score is defined as the average of the three scores. In the following subsections, we elaborate on them in detail one by one.

\subsubsection{\textbf{Text-based Global-Local interaction Unit}}
Text is the basic but imperative information for entity linking. Previous methods utilized the hidden status of \text{[CLS]} as global features~\cite{DBLP:conf/emnlp/WuPJRZ20} while losing the local features, or integrated Conv1D to measure character level similarity whereas ignoring the global coherence. To measure global consistency, we use the dot product of two normalized global features as the global-to-global score, mathematically formulated as,
\begin{equation}
    \mathcal{S}_T^{G2G} = \mathbf{t}_\mathbf{E}^G \cdot \mathbf{t}_\mathbf{M}^G.
    \label{eq:s_t^g2g}
\end{equation}
Based on the designed unified textual input, Equation~\ref{eq:s_t^g2g} directly measures the global correlation of text input of mention and entity. Then we make further efforts to discover fine-grained clues among local features. Specifically, we utilize the attention mechanism to capture the context of different local features, and the representation is calculated as follows:
\begin{equation}
    \begin{split}
        Q, K, V  &= \mathbf{T}_\mathbf{E}^L \boldsymbol{W}_{tq}, \mathbf{T}_\mathbf{M}^L \boldsymbol{W}_{tk}, \mathbf{T}_\mathbf{M}^L \boldsymbol{W}_{tv}, \\
        H_t &= \text{softmax}(\frac{QK^T}{\sqrt{d_T}})V,
    \end{split}
\end{equation}
where $\boldsymbol{W}_{tq}$, $\boldsymbol{W}_{tk}$,  $\boldsymbol{W}_{tv} \in \mathbb{R}^{d_T \times d_t}$ are learnable matrices, and $d_t$ represents the dimension inside TGLU. Then we adopt mean pooling and layer norm over $H_t$ to get the context vector, and further measure the global-to-local score between the vector and the projected $\mathbf{t}_{\mathbf{E}}^L$ as follows:
\begin{equation}
    \begin{gathered}
        h_t = \text{LayerNorm} \left( \text{MeanPooling} \left( H_t \right)  \right), \\
        \mathcal{S}_T^{G2L} = \text{FC} \left( \mathbf{t}_{\mathbf{E}}^G \right) \cdot h_t,
    \end{gathered}    
\end{equation}
where the fully connected (FC) layer consists of  $\boldsymbol{W}_{t1} \in \mathbb{R}^{d_T \times d_t}$ and $\boldsymbol{b}_{t1} \in \mathbb{R}^{d_t}$. Afterwards, the matching score of TGLU is defined as the average of $\mathcal{S}_T^{G2G}$ and $\mathcal{S}_T^{G2L}$ following Equation \ref{eq:ST}.

\subsubsection{\textbf{Vision-based DuaL interaction Unit}}
Visual information plays an essential role in multimodal entity linking because images directly depict entities or a scene of the related object, which reflects explicit indication. However, the noise in images brings difficulties for MEL and further impairs performance. To overcome this issue, we propose VDLU with a dual-gated mechanism. Different from threshold filter~\cite{DBLP:conf/dasfaa/ZhangLY21}, the dual-gated mechanism considers feature interaction from both mention's view and entity's view to resist noise, where the gate is designed to control the feature interaction. From an overview, the VDLU can be formulated as:
\begin{equation}
    \begin{split}
        \mathcal{S}_V^{E2M} &= \text{DUAL}_{E2M} \left ( \mathbf{v}_\mathbf{E}^G, \mathbf{v}_\mathbf{M}^G, \mathbf{V}_\mathbf{M}^L \right), \\
        \mathcal{S}_V^{M2E} &= \text{DUAL}_{M2E} \left( \mathbf{v}_\mathbf{M}^G, \mathbf{v}_\mathbf{E}^G, \mathbf{V}_\mathbf{E}^L \right),
    \end{split}
\end{equation}
where $\text{DUAL}_{A2B}(\mathbf{v}_\mathbf{A}^G, \mathbf{v}_\mathbf{B}^G, \mathbf{V}_\mathbf{B}^L)$ represents the dual-gated mechanism by considering the feature interaction from $A$ to $B$. Without losing generality, here we use $A$ and $B$ to represent entity ($E$) or mention ($M$) for illustrating $\text{DUAL}_{A2B}(\cdot,\cdot,\cdot)$ function. We first utilize mean pooling and layer norm over $\mathbf{V}_\mathbf{B}^L$ to get the pooled vector $\bar{h}_p$ and combine it with $\mathbf{v}_\mathbf{A}^G$ as follows:
\begin{equation}
    \begin{gathered}
        \bar{h}_p = \text{MeanPooling} \left( \mathbf{V}_\mathbf{B}^L \right), \\ 
        h_{vc} = \text{FC} \left( \text{LayerNorm} \left( \bar{h}_p + \mathbf{v}_\mathbf{A}^G \right) \right) ,
    \end{gathered}
    \label{eq:VDLU_eq1}
\end{equation}
where the FC layer contains trainable parameters $\boldsymbol{W}_{v1} \mathbb{R}^{d_v \times d_v}$ and $\boldsymbol{b}_{v1} \in \mathbb{R}^{d_v}$. After that, we obtain the gate value by another FC layer connected with an activation function, which is applied to control the feature interaction with the fused feature $\mathbf{v}_\mathbf{B}^G$, i.e., 
\begin{equation}
    \begin{gathered}
        h_{vg} = \text{Tanh} \left( \text{FC} \left( h_{vc} \right) \right), \\
        h_{v} = \text{LayerNorm} \left( h_{vg} * h_{vc} +  \mathbf{v}_\mathbf{B}^G\right), 
    \end{gathered}
    \label{eq:VDLU_eq2}
\end{equation}
where the gate FC layer includes $\boldsymbol{W}_{v2} \in \mathbb{R}^{d_v \times 1}$ and $\boldsymbol{b}_{v2} \in \mathbb{R}$ and converts $h_{vc}$ into a real number. Thus, three input features are sufficiently interacted and fused. Afterwards, the score of $\mathcal{S}_V^{A2B}$ is calculated by the dot product between $h_{v}$ and $\mathbf{v}_\mathbf{A}^G$:
\begin{equation}
    \mathcal{S}_V^{A2B} = h_{v} \cdot \mathbf{v}_\mathbf{A}^G.
    \label{eq:VDLU_eq3}
\end{equation}
According to the above formulas Equation \ref{eq:VDLU_eq1} - Equation \ref{eq:VDLU_eq3} on the calculation of $\text{DUAL}_{A2B}$, similarly, we can obtain $\mathcal{S}_{V}^{E2M}$ and $\mathcal{S}_{V}^{M2E}$, which lead us to the final score $\mathcal{S}_V$.

\subsubsection{\textbf{Cross-Modal Fusion-based interaction Unit}}
As mentioned before, the images contain implicit indications which can be inferred from multiple modalities. To highlight the subtle cues or signals among different features, the designed CMFU considers the cross-modal alignment and fusion via a gated function based on the extracted local and global features. In order to obtain the unit-related features for the subsequent operations as well as compact the dimension of features, we convert textual and visual features via two fully connected layers as follows,
\begin{equation}
    \begin{split}
        h_{et}, h_{mt} &= \text{FC}_{c1} \left( \mathbf{t}_\mathbf{E}^G \right), \text{FC}_{c1} \left( \mathbf{t}_\mathbf{M}^G \right), \\
        H_{ev}, H_{mv} &= \text{FC}_{c2} \left( \mathbf{V}_\mathbf{E}^L \right), \text{FC}_{c2} \left( \mathbf{V}_\mathbf{M}^L \right),
    \end{split}
    \label{eq:CMFU_eq1}
\end{equation}
in which $\text{FC}_{c1}$ is defined by $\boldsymbol{W}_{c1} \in \mathbb{R}^{d_T \times d_c}$ and $\boldsymbol{b}_{c1} \in \mathbb{R}^{d_c}$,  $\text{FC}_{c2}$ is defined by $\boldsymbol{W}_{c2} \in \mathbb{R}^{d_v \times d_c}$ and $\boldsymbol{b}_{c2} \in \mathbb{R}^{d_c}$. After projection, we introduce a function  $\text{FUSE}(h_{ot}, H_{ov})$  for the fine-grained fusion of textual and visual features, where $o$ represents entity ($e$) or mention ($m$). Without losing generality, we take the fusion of entity side as an example. First, the element-wise dot product scores of textual and visual features are applied to guide the aggregation of image patch information,
\begin{equation}
    \begin{gathered}
        \alpha_{i} = \frac{\text{exp} \left( h_{et} \cdot H_{ev}^i \right)}{\sum_{i}^{n+1} \text{exp} \left( h_{et} \cdot H_{ev}^i \right)},\\
        h_{ec} = \sum_{i}^{n+1} \alpha_{i} * H_{ev}^i, i \in \left[ 1, 2, \dots, (n+1) \right].
    \end{gathered}
    \label{eq:CMFU_eq2}
\end{equation}
Meanwhile, the intensity of textual information is evaluated with a gate operation,
\begin{equation}
    h_{eg} = \text{Tanh} \left( \text{FC}_{c3} \left( h_{et} \right) \right),
    \label{eq:CMFU_eq3}
\end{equation}
where $\text{FC}_{c3}$ is composed of a learnable matrix $\boldsymbol{W}_{c3} \in \mathbb{R}^{d_c \times d_c}$ and a learnable bias vector $\boldsymbol{b}_{c3} \in \mathbb{R}^{d_c}$. Based on the gate value, the entity context is summarized by,
\begin{equation}
    h_e = \text{LayerNorm} \left( h_{eg} * h_{et} +  h_{ec} \right).
    \label{eq:CMFU_eq4}
\end{equation}
Following the operations Equation \ref{eq:CMFU_eq2} - Equation \ref{eq:CMFU_eq4} by replacing inputs $h_{et}$ and $H_{ev}$ with $h_{mt}$ and $H_{mv}$, we can also get the mention-side context vector $h_m$.  
Then, the score is calculated by the dot product, 
\begin{equation}
    \mathcal{S}_C = h_e \cdot h_m.
\end{equation}

\subsection{Unit-Consistent Objective Function}
Based on the score that we calculate above, we jointly train both the encoding layer and the interaction layer with a contrastive training loss function. Hence, the model learns to rate the positive mention-entity pairs higher and the negative mention-entity pairs lower. This loss function can be formulated as
\begin{equation}
    \mathcal{L}_{O} = -\log \frac{\exp \left(\mathcal{U}(\textbf{M}, \textbf{E}) \right)}{\sum_i \exp \left( \mathcal{U}(\textbf{M}, \textbf{E}_i^\prime) \right)},
\end{equation}
where $\mathbf{E}^\prime_i$ is the negative entity from the knowledge base $\mathcal{E}$ and we use in-batch negative sampling in our implementation. However, the function $\mathcal{U}(\mathbf{M}, \mathbf{E})$ calculates the average scores of three units. This may result in one of the units taking the dominant position,  causing the whole model to excessively rely on its score. In addition, inconsistencies in scoring may also occur as different units consider different perspectives. To this end, we propose to design independent loss functions for each unit as follows,
\begin{equation}
    \mathcal{L}_{X} = -\log \frac{\exp \left(\mathcal{U}_{X}(\textbf{M}, \textbf{E}) \right)}{\sum_i \exp \left( \mathcal{U}_{X}(\textbf{M}, \textbf{E}_i^\prime) \right)}, X \in \{T, V, C\},
\end{equation}
where $X$ represents any interaction units. Eventually, the optimization objective function is 
\begin{equation}
    \mathcal{L} = \mathcal{L}_O + \underbrace{\mathcal{L}_T + \mathcal{L}_V + \mathcal{L}_C}_{\text{unit-consistent loss function}}.
    \label{eq:all_loss}
\end{equation}
As for the evaluation stage, we use $\mathcal{S}=\mathcal{U}(\mathbf{M}, \mathbf{E})$, i.e., the average scores of three interaction units, as the evidence for ranking entities.

\section{Experiments}
In this section, we carried out comprehensive experiments on three public multimodal entity linking datasets to sufficiently validate the effectiveness of our proposed MIMIC. We are intended to investigate the following research questions (RQ):

\begin{itemize}[leftmargin=*]
  \item \textbf{RQ1.} How does the proposed MIMIC perform compared with various baselines? 
  \item \textbf{RQ2.} How do the generalization abilities of MIMIC and other baselines perform in low-resource scenarios?
  \item \textbf{RQ3.} How do the three proposed interaction units and unit-consistent objective function affect performance?
  \item \textbf{RQ4.} How does the model performance change with the parameters?
\end{itemize}

\subsection{Experimental Setup}

\subsubsection{Datasets}
In the experiments, we selected three public MEL datasets \textbf{WikiMEL}, \textbf{RichpediaMEL}~\cite{DBLP:conf/sigir/WangWC22} and \textbf{WikiDiverse}~\cite{DBLP:conf/acl/WangTGLWYCX22} to verify the effectiveness of our proposed method.  

\textbf{WikiMEL}~\cite{DBLP:conf/sigir/WangWC22} is collected from Wikipedia entities pages and contains more than 22k multimodal sentences. \textbf{RichpediaMEL}~\cite{DBLP:conf/sigir/WangWC22} is obtained form a MMKG Richpedia~\cite{DBLP:journals/bdr/WangWQZ20}. The authors of RichpediaMEL first extracted entities form Richpedia and then obtain multimodal information form Wikidata~\cite{DBLP:journals/cacm/VrandecicK14}. The main entity types of WikiMEL and RichpedaiMEl are person. \textbf{WikiDiverse}~\cite{DBLP:conf/acl/WangTGLWYCX22} is constructed from Wikinews and covers various topics including sports, technology, economy and so on. We used Wikidata as our knowledge base (KB) and removed the mention that we could not find the corresponding entity in Wikidata. Linking a mention to a large-scale MMKG or multimodal knowledge base is extremely time-consuming, especially when taking images into consideration. To fairly conduct experiments, we followed the previous studies~\cite{DBLP:conf/sigir/WangWC22}, and used a subset KB of Wikidata for each dataset. We used the original split of the three datasets. For both WikiMEL and RichpediaMEL, 70\%, 10\% and 20\% of the data are divided into training set, validation set and test set respectively. As for WikiDiverses, the proportions are 80\%, 10\% and 10\%. 
Appendix~\ref{sec:details_of_datasets} provides detailed statistical information about the datasets.

\subsubsection{Baselines} We compared our method with various competitive baselines including text-based methods, MEL methods and Vision-and-Language Pre-training (VLP) models. Specifically, the text-based methods include \textbf{BLINK}~\cite{DBLP:conf/emnlp/WuPJRZ20}, \textbf{BERT}~\cite{DBLP:conf/naacl/DevlinCLT19}, \textbf{RoBERTa}~\cite{DBLP:journals/corr/abs-1907-11692}. MEL methods contain \textbf{DZMNED}~\cite{DBLP:conf/acl/CarvalhoMN18}, \textbf{JMEL}~\cite{DBLP:conf/ecir/AdjaliBFBG20}, \textbf{VELML}~\cite{DBLP:journals/dint/ZhengWWQ22}, \textbf{GHMFC}~\cite{DBLP:conf/sigir/WangWC22}. Moreover,  the VLP models include \textbf{CLIP}~\cite{DBLP:conf/icml/RadfordKHRGASAM21},  \textbf{ViLT}~\cite{DBLP:conf/icml/KimSK21}, \textbf{ALBEF}~\cite{DBLP:conf/nips/LiSGJXH21}, \textbf{METER}~\cite{DBLP:conf/cvpr/DouXGWWWZZYP0022},   and these models are usually pre-trained with large-scale image-text corpus with image-text matching loss and mask language modeling loss. Detailed descriptions of baselines are provided in Appendix~\ref{sec:desc_of_baselines}.

\subsubsection{Evaluation Metrics} When evaluating, we calculated the similarity between a mention and all entities of KB to measure their aligning probability. The similarity scores are sorted in descending order to calculate \textbf{H@k}, \textbf{MRR} and \textbf{MR}. We provide the calculation methods for each metric in Appendix~\ref{sec:eval_metric}.

H@k indicates the hit rate of the ground truth entity when only considering the top-k ranked entities. MRR represents the mean reciprocal rank of the ground truth entity. MR is the mean rank of the ground truth entity among all entities. Hence, both H@k and MRR are the higher the better, but a lower MR indicates better performance.

\begin{table*}[ht]
    \caption{Performance comparison on three MEL datasets. We run each method three times with different random seeds and report the mean value of every metric. The best score is highlighted in \textbf{bold} and the second best score is \underline{underlined}. The symbol "$\smallstar$" denotes the p-value of the t-test compared with the second best score is lower than 0.005 and "$\ast$" means the p-value is lower than 0.01 but higher than 0.005.}
    \label{tab:main}
    \resizebox{\textwidth}{!}{
    \begin{tabular}{c|ccccc|ccccc|ccccc}
    \toprule
    \multirow{2}{*}{Model} & \multicolumn{5}{c|}{WikiMEL}                                                       & \multicolumn{5}{c|}{RichpediaMEL}                                                  & \multicolumn{5}{c}{WikiDiverse}                                                     \\ \cmidrule(lr){2-6}\cmidrule(lr){7-11}\cmidrule(lr){12-16}
                                                        & H@1↑           & H@3↑           & H@5↑           & MRR↑           & MR↓            & H@1↑           & H@3↑           & H@5↑           & MRR↑           & MR↓            & H@1↑           & H@3↑           & H@5↑           & MRR↑           & MR↓             \\ \midrule
    BLINK~\cite{DBLP:conf/emnlp/WuPJRZ20}               & 74.66          & 86.63          & 90.57          & 81.72          & 51.48          & 58.47          & 81.51          & 88.09          & 71.39          & 178.57         & 57.14          & 78.04          & 85.32          & 69.15          & 332.03          \\
    BERT~\cite{DBLP:conf/naacl/DevlinCLT19}             & 74.82          & 86.79          & 90.47          & 81.78          & 51.23          & 59.55          & 81.12          & 87.16          & 71.67          & 278.08         & 55.77          & 75.73          & 83.11          & 67.38          & 373.96          \\
    RoBERTa~\cite{DBLP:journals/corr/abs-1907-11692}    & 73.75          & 85.85          & 89.80          & 80.86          & 31.02          & 61.34          & 81.56          & 87.15          & 72.80          & 218.16         & 59.46          & 78.54          & 85.08          & 70.52          & 405.22          \\ \midrule
    DZMNED~\cite{DBLP:conf/acl/CarvalhoMN18}            & 78.82          & 90.02          & 92.62          & 84.97          & 152.58         & 68.16          & 82.94          & 87.33          & 76.63          & 313.85         & 56.90          & 75.34          & 81.41          & 67.59          & 563.26          \\
    JMEL~\cite{DBLP:conf/ecir/AdjaliBFBG20}             & 64.65          & 79.99          & 84.34          & 73.39          & 285.14         & 48.82          & 66.77          & 73.99          & 60.06          & 470.90         & 37.38          & 54.23          & 61.00          & 48.19          & 996.63          \\
    VELML~\cite{DBLP:journals/dint/ZhengWWQ22}          & 76.62          & 88.75          & 91.96          & 83.42          & 102.72         & 67.71          & 84.57          & 89.17          & 77.19          & 332.85         & 54.56          & 74.43          & 81.15          & 66.13          & 463.25          \\
    GHMFC~\cite{DBLP:conf/sigir/WangWC22}               & 76.55          & 88.40          & 92.01          & 83.36          & 54.75          & {\ul 72.92}    & {\ul 86.85}    & {\ul 90.60}    & {\ul 80.76}    & 214.64         & 60.27          & 79.40          & 84.74          & 70.99          & 628.87          \\ \midrule
    CLIP~\cite{DBLP:conf/icml/RadfordKHRGASAM21}        & {\ul 83.23}    & {\ul 92.10}    & {\ul 94.51}    & {\ul 88.23}    & {\ul 17.60}    & 67.78          & 85.22          & 90.04          & 77.57          & {\ul 107.16}   & {\ul 61.21}    & {\ul 79.63}    & {\ul 85.18}    & {\ul 71.69}    & 313.35          \\
    ViLT~\cite{DBLP:conf/icml/KimSK21}                  & 72.64          & 84.51          & 87.86          & 79.46          & 220.76         & 45.85          & 62.96          & 69.80          & 56.63          & 675.93         & 34.39          & 51.07          & 57.83          & 45.22          & 2421.49         \\
    ALBEF~\cite{DBLP:conf/nips/LiSGJXH21}               & 78.64          & 88.93          & 91.75          & 84.56          & 47.95          & 65.17          & 82.84          & 88.28          & 75.29          & 122.30         & 60.59          & 75.59          & 81.30          & 69.93          & {\ul 291.17}    \\
    METER~\cite{DBLP:conf/cvpr/DouXGWWWZZYP0022}        & 72.46          & 84.41          & 88.17          & 79.49          & 111.90         & 63.96          & 82.24          & 87.08          & 74.15          & 376.42         & 53.14          & 70.93          & 77.59          & 63.71          & 944.48          \\ \midrule
    MIMIC                  & $\textbf{87.98}^\smallstar$ & $\textbf{95.07}^\ast$ & $\textbf{96.37}^\ast$ & $\textbf{91.82}^\smallstar$ & $\textbf{11.02}$ & $\textbf{81.02}^\smallstar$ & $\textbf{91.77}^\smallstar$ & $\textbf{94.38}^\smallstar$ & $\textbf{86.95}^\smallstar$ & $\textbf{55.11}^\smallstar$ & $\textbf{63.51}^\ast$ & $\textbf{81.04}$ & $\textbf{86.43}^\ast$ & $\textbf{73.44}^\ast$ & $\textbf{227.08}$ \\ \bottomrule 
    \end{tabular}}
\end{table*}

\subsubsection{Implementation Details} Our model weights are initialized with pre-trained CLIP-Vit-Base-Patch32\footnote{https://huggingface.co/openai/clip-vit-base-patch32}, where ViT-B/32 Transformer architecture is employed as an image encoder and the patch size P is 32. All images are rescaled into 224 $\times$ 224 resolution and we used zero padding to handle the mentions and entities without images. The maximal length of text input is set to 40 and the dimension of textual output features, i.e., $d_T$ is set to 512. As for the parameters in the interaction layer, $d_t$, $d_v$ and $d_c$ are set to 96 for all three datasets.  We used the deep learning framework PyTorch~\cite{DBLP:conf/nips/PaszkeGMLBCKLGA19} to implement our method and trained it on a device equipped with an Intel(R) Xeon(R) Gold 6248R CPU and a GeForce RTX 3090 GPU. We trained our MIMIC using AdamW~\cite{DBLP:conf/iclr/LoshchilovH19} optimizer with a batch size of 128 to accommodate maximal GPU memory and betas are set to (0.9, 0.999). The number of epochs and learning rate are well-tuned to 20 and $1 \times 10^{-5}$ respectively. All methods are evaluated on the validation set and the checkpoint with the highest MRR is selected to evaluate on the test set. As for the baselines, we re-implemented DZMNED, JMEL, VELML according to the original literature due to they did not release the code. We ran the official implementations of the other baselines with their default settings.

\subsection{Experimental Results}
\subsubsection{Overall Comparison (RQ1).}
We compared our proposed MIMIC with baselines on three benchmark datasets. As shown in Table \ref{tab:main}, average scores of the performance on the test set across three random runs are reported. Overall, our proposed MIMIC achieves the best metrics on three datasets, with 3.59\%, 6.19\%, 1.75\% absolute improvement of MRR on WikiMEL, RichpediaMEL and WikiDiverse respectively. This demonstrates the superiority of MIMIC for solving the MEL task. According to the experimental results of Table \ref{tab:main}, we further have the following  observations and analysis.

First, compared with MEL and VLP methods, the text-based approaches show promising performance. It suggests that textual information is still the basic but crucial modality for MEL because the text provides a measurement from the surface. It is noticed that BLINK slightly underperforms BERT on WikiMEL and RichpediaMEL but outperforms BERT on WikiDiverse. Although BLINK utilizes two encoders to extract global representations for mentions and entities separately, similar to BERT, it ignores the local features in the short and abbreviated text which impairs their performance  Moreover, compared with the state-of-the-art MEL methods, the text-based approaches still have a gap in performance because they only rely on textual inputs but ignore visual information, which brings difficulty to identify vague mentions within the limited text.

Second, different MEL methods have their respective pros and cons. Benefiting from the hierarchical fine-grained co-attention mechanism, GHMFC achieves the best result on three datasets among all MEL baselines. In particular, compared with all other baselines, GHMFC achieves 72.92\% and 80.76\% for H@1 and MRR respectively on RichpediaMEL, which is only inferior to our proposed MIMIC. It indicates that effectively incorporating visual features into multimodal interaction contributes to improving the performance of MEL. Different MEL methods show a large gap. As shown in Table~\ref{tab:main}, JMEL underperforms DZMEND, VELML and GHMFC on three datasets, which may result from the strategy of multimodal fusion. JMEL utilizes simple concatenation and a fully connected layer to fuse textual and visual features. In contrast, both DZMEND and VELML use additional attention mechanism to fuse different features. It suggests that shallow modality interaction and naive multimodal fusion bring no improvement even degeneration on the performance of MEL. 

Third, VLP methods also demonstrate competitive evaluation results compared with MEL baselines. CLIP achieves the second best metrics except for MR on both WikiMEL and WikiDiverse, which benefits from pre-training with the large-scale image-text corpus. ALBEF and METER also display similar results with CLIP. We argue that these methods could be further exploited by considering fine-grained interaction and delicate designed fusion.

Finally, the experimental results demonstrate the effectiveness and superiority of our proposed MIMIC. Compared with the second best metric, MIMIC gains 4.75\%, 8.1\% and 2.3\% absolute improvement of Hit@1 on WikiMEL, RichpediaMEL and WikiDiverse respectively. We also performed significant tests to further validate the statistical evidence between MIMIC and other baselines. Specifically, the p-values of MRR on three datasets are 0.002, 0.0001 and 0.009 respectively. All p-values are under 0.01 and show a significant advantage in statistics.

\begin{figure*}[ht]
    \centering
    \includegraphics[width=0.96\textwidth]{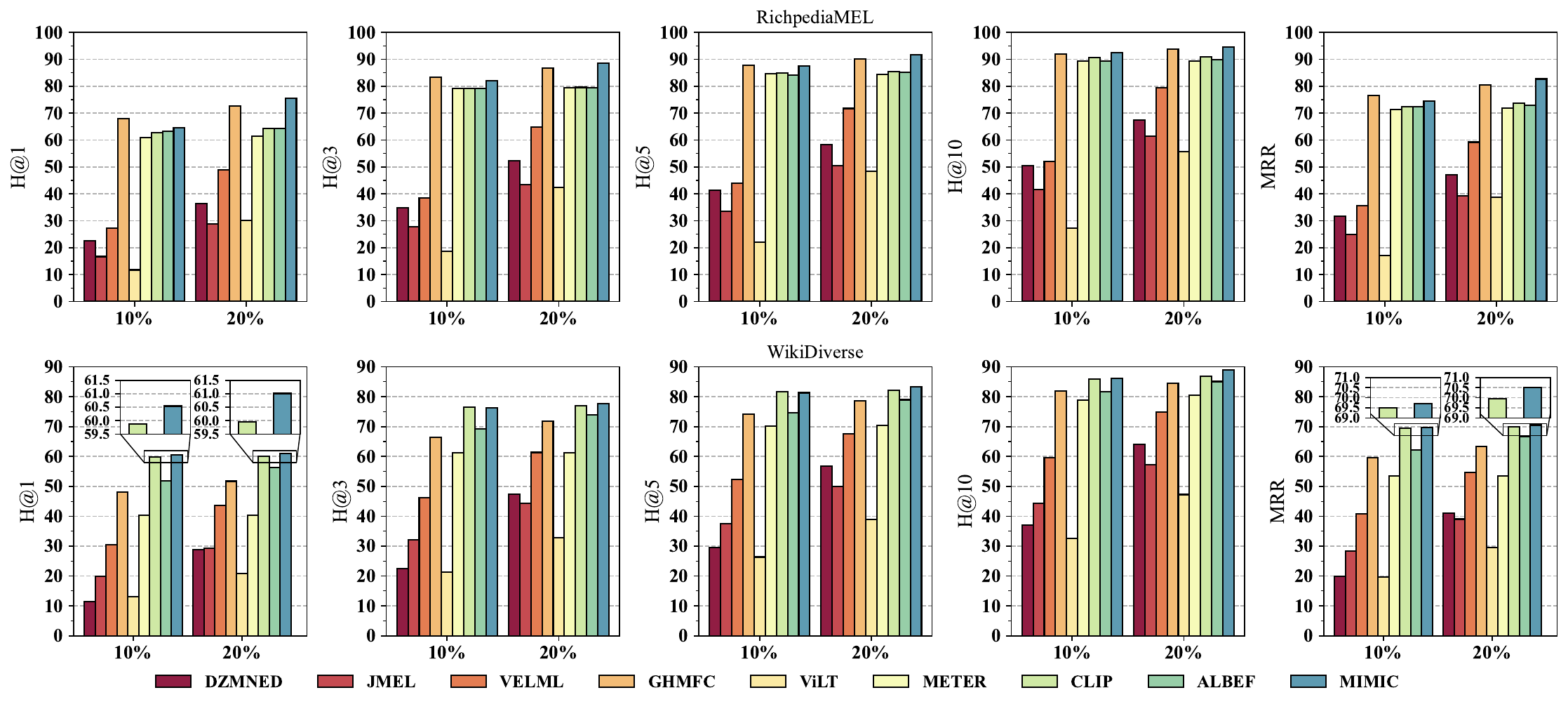}
    \caption{Performance comparison of low resource settings on RichpediaMEL and WikiDiverse. Details are zoomed in for better visualization.}
    \label{fig:low_res}
\end{figure*}

\subsubsection{Low resource setting (RQ2).}
Collecting and acquiring high-quality annotated data is extremely laborious and time-consuming. Therefore, it is necessary to investigate the performance of the models in low-resource scenarios. We conducted experiments using 10\% and 20\% of the training data while keeping the validation and test sets unchanged. Experimental results are shown in Figure~\ref{fig:low_res}. In overview, most of the MEL methods manifest a significant drop in performance. Except for ViLT, other VLP methods benefit from large-scale multimodal pre-training and show a slight decrease in performance, which means that well-trained weights guarantee a reasonable performance in a low resource setting. With the increase in training data, nearly all methods e.g., DZMNED, JMEL and VELML, show an obvious improvement, which means sufficient training data is necessary to improve the performance. Notably, GHMFC outperforms our proposed MIMIC with 10\% training data on RichpediaMEL but underperforms MIMIC with 10\% training data on WikiDiverse while showing a clear gap. It suggests that GHMFC does not generalize well on different datasets. 
When the proportion comes to 20\%, our proposed MIMIC surpasses GHMFC in every metric on RichpediaMEL and shows an obvious margin. From 10\% to 20\%, the absolute improvement of H@1, H@3 and MRR of MIMIC are 11.2\%, 6.6\% and 8.11\%, respectively. This phenomenon reveals that detailed inter-modal and intra-modal interaction units of MIMIC have better adaptability with the increase in training data. As for WikiDiverse, CLIP slightly underperforms MIMIC on H@1 and MRR in the 10\% setting. With the increase in training proportion, the gap between MIMIC and CLIP gradually becomes larger, which validates MIMIC has better capability and potential in the low resources scenario.

\begin{table*}[ht]
\caption{Experimental results of ablation studies. The best scores are highlighted in bold.}
\label{tab:ablation}
\begin{tabular}{c|cccccc|cccccc}
\toprule
\multirow{2}{*}{Model}      & \multicolumn{6}{c|}{WikiMEL}                  & \multicolumn{6}{c}{RichpediaMEL}              \\ \cmidrule(lr){2-7}\cmidrule(lr){8-13}
                            & H@1↑  & H@3↑  & H@5↑  & H@10↑ & H@20↑ & MRR↑  & H@1↑  & H@3↑  & H@5↑  & H@10↑ & H@20↑ & MRR↑  \\ \midrule
MIMIC                       & \textbf{87.98} &\textbf{ 95.07} & \textbf{96.37} & 97.80 & 98.73 & \textbf{91.82} & \textbf{81.02} & \textbf{91.77} & \textbf{94.38} & \textbf{96.69} & \textbf{98.04} & \textbf{86.95} \\ \midrule
w/o $\mathcal{L}_T$         & 86.13 & 93.69 & 95.74 & 97.66 & 98.57 & 90.42 & 72.82 & 89.05 & 93.12 & 96.15 & 97.61 & 81.61 \\
w/o $\mathcal{L}_V$         & 86.71 & 94.43 & 96.25 & \textbf{98.01} & \textbf{98.80} & 90.94 & 78.72 & 90.23 & 93.66 & 96.04 & 97.61 & 85.15 \\
w/o $\mathcal{L}_C$         & 86.67 & 94.04 & 95.69 & 97.21 & 98.18 & 90.74 & 79.65 & 89.89 & 92.56 & 94.92 & 96.94 & 85.38 \\ \midrule
w/o TGLU + $\mathcal{L}_T$  & 85.03 & 92.36 & 94.35 & 95.94 & 97.27 & 89.18 & 74.48 & 85.37 & 88.71 & 92.00 & 94.02 & 80.74 \\
w/o VDLU + $\mathcal{L}_V$  & 83.46 & 93.33 & 95.47 & 97.23 & 98.18 & 88.74 & 74.12 & 89.47 & 92.81 & 95.82 & 97.61 & 82.37 \\
w/o CMFU + $\mathcal{L}_C$  & 84.60 & 92.90 & 94.82 & 96.42 & 97.35 & 89.14 & 76.98 & 88.29 & 91.30 & 94.22 & 96.15 & 83.39 \\ \bottomrule
\end{tabular}
\end{table*}

\begin{figure*}[ht]
    \centering
    \includegraphics[width=0.95\textwidth]{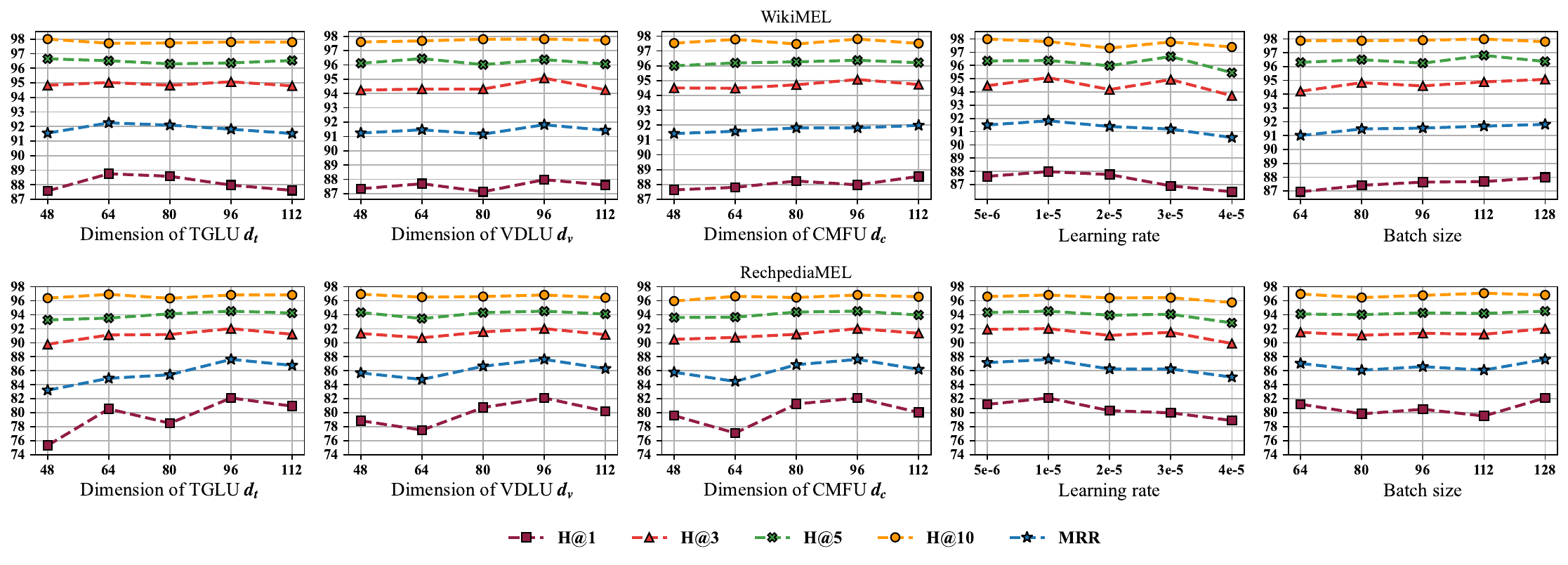}
    \caption{Parameter sensitivity analysis on WikiMEL and RichpediaMEL regarding different values.}
    \label{fig:param}
\end{figure*}

\subsubsection{Ablation Study (RQ3).}
To delve into the effect of three proposed interaction units and unit-consistent loss function, we designed two groups of experiments for the ablation study. In the first group, we remove $\mathcal{L}_T$, $\mathcal{L}_V$ and $\mathcal{L}_C$ separately from loss function, i.e., Equation \ref{eq:all_loss}. We denote these variants as w/o $\mathcal{L}_T$, w/o $\mathcal{L}_V$ and w/o $\mathcal{L}_C$ respectively. In the second group, we further compare MIMIC with the following variants: (1) w/o TGLU + $\mathcal{L}_T$: removing the text-based global-local matching unit and its loss function; (2) w/o VDLU + $\mathcal{L}_V$: removing the vision-based dual matching units along with its loss function; (3) w/o CMFU + $\mathcal{L}_C$: removing the cross-modal fusion-based matching unit and its loss function. Table~\ref{tab:ablation} illustrates the experimental results.

Overall, removing any interaction unit or loss function from the full model results in an evident decline in almost every metric to varying degrees, which proves the effectiveness of the designed interaction units and unit-consistent loss function. The performance of w/o $\mathcal{L}_T$ and w/o $\mathcal{L}_C$ drops marginally on WikiMEL. It is noticed that w/o $\mathcal{L}_V$ outperforms the full model diminutively on H@10 and H@20. However, the model w/o $\mathcal{L}_V$ shows an obvious decline in H@1 and MRR. One possible reason is that $\mathcal{L}_V$ improves overall performance but has a side effect on some hard samples depending on the dataset. On RichpediaMEL, a significant performance drop of w/o $\mathcal{L}_T$ can be observed. H@1 degrades from 81.02\% to 72.82\% and MRR drops from 86.95\% to 81.61\%. This demonstrates the unit-consistent loss function improves intra-modal and inter-modal learning because it helps that the ground truth entity could be retrieved from any single interaction unit. The unit-consistent loss function also alleviates the modality inconsistency caused by noisy data. Moreover, excluding any interaction units leads to a decrease in performance as well. Specifically, the variant w/o VDLU + $\mathcal{L}_V$ shows the worst H@1 and MRR on WikiMEL. In terms of RichpediaMEL, the model w/o TGLU + $\mathcal{L}_T$ has the worst MRR, which suggests that the two datasets have different salient modalities and schemata. Hence it is necessary to explore the interaction and fusion in multimodal and multi-grained ways. The combination of our proposed interaction matching units gives an effective boost to most metrics, proving the efficacy of our design.

\subsubsection{Parameter Sensitivity Analysis (RQ4).}

In this section, we investigated the sensitivity of parameters on two datasets, WikiMEL and RichpediaMEL. The experimental results are shown in Figure~\ref{fig:param}. First, we analyzed the effect of various dimensions of TGLU, VDLU and CMFU, namely $d_t$, $d_v$ and $d_c$. We can see that the performance raise up gradually with the increase in dimension and then drops slowly. It suggests that three interaction units need a proper dimension to encode semantics features, but a large dimension may cause redundancy, leading to a decrease in performance. Second, we explored the impact of the learning rate. The result shows that performance benefits from a small and suitable learning rate because we initialized MIMIC with pre-trained model weights. As the learning rate gets larger, the performance starts to degenerate because of converging to a suboptimal solution. We also analyzed the effect of batch size. Based on the results, a larger batch size generally improves the performance of MIMIC. The reason is that MIMIC utilizes in-batch contrastive learning. Hence a large batch size means more negative samples in a single batch, which could enhance the representation learning process.

\section{Conclusion}
In this paper, we proposed a novel Multi-Grained Multimodal Interaction network (MIMIC) for solving multimodal entity linking task, which comprehensively explores intra-modal and inter-modal patterns to extract explicit and implicit clues. Concretely, we first designed a unified input format to encode both entities and mentions into the same vector space, which reduces the feature gap between entities and mentions. Then, we devised three interaction units, namely Text-based Global-Local interaction Unit, Vision-based DuaL interaction Unit and Cross-Modal Fusion-based interaction Unit, to explore the explicit and implicit semantics relevance within extracted multimodal features. Afterwards, we also introduced a unit-consistent loss function to improve multimodal learning and enhance the consistency of our model against noisy data. Extensive experiments on three public datasets have validated the effectiveness of our MIMIC framework compared with several state-of-the-art baseline methods.

\begin{acks}
This work was supported by the grants from National Natural Science Foundation of China (No.U22B2059, 62222213, 62276245, 62072423), the Anhui Provincial Natural Science Foundation (Grant No. 2008085J31), and the USTC Research Funds of the Double First-Class Initiative (No.YD2150002009).
\end{acks}

\bibliographystyle{plain}
\balance
\bibliography{bibtex}

\begin{thebibliography}{10}

\bibitem{DBLP:conf/ecir/AdjaliBFBG20}
Omar Adjali, Romaric Besan{\c{c}}on, Olivier Ferret, Herv{\'{e}}~Le Borgne, and
  Brigitte Grau.
\newblock Multimodal entity linking for tweets.
\newblock In {\em {ECIR} {(1)}}, volume 12035 of {\em Lecture Notes in Computer
  Science}, pages 463--478. Springer, 2020.

\bibitem{DBLP:conf/cikm/AhmadvandSCA19}
Ali Ahmadvand, Harshita Sahijwani, Jason~Ingyu Choi, and Eugene Agichtein.
\newblock Concet: Entity-aware topic classification for open-domain
  conversational agents.
\newblock In {\em {CIKM}}, pages 1371--1380. {ACM}, 2019.

\bibitem{DBLP:conf/iclr/CaoI0P21}
Nicola~De Cao, Gautier Izacard, Sebastian Riedel, and Fabio Petroni.
\newblock Autoregressive entity retrieval.
\newblock In {\em {ICLR}}. OpenReview.net, 2021.

\bibitem{DBLP:conf/coling/0002HLL18}
Yixin Cao, Lei Hou, Juanzi Li, and Zhiyuan Liu.
\newblock Neural collective entity linking.
\newblock In {\em {COLING}}, pages 675--686. Association for Computational
  Linguistics, 2018.

\bibitem{DBLP:conf/acl/CaoHJCL17}
Yixin Cao, Lifu Huang, Heng Ji, Xu~Chen, and Juanzi Li.
\newblock Bridge text and knowledge by learning multi-prototype entity mention
  embedding.
\newblock In {\em {ACL} {(1)}}, pages 1623--1633. Association for Computational
  Linguistics, 2017.

\bibitem{DBLP:conf/mm/ChenDLZH19}
Hui Chen, Guiguang Ding, Zijia Lin, Sicheng Zhao, and Jungong Han.
\newblock Cross-modal image-text retrieval with semantic consistency.
\newblock In {\em {ACM} Multimedia}, pages 1749--1757. {ACM}, 2019.

\bibitem{DBLP:conf/kdd/ChenL00WYC22}
Liyi Chen, Zhi Li, Tong Xu, Han Wu, Zhefeng Wang, Nicholas~Jing Yuan, and
  Enhong Chen.
\newblock Multi-modal siamese network for entity alignment.
\newblock In {\em {KDD}}, pages 118--126. {ACM}, 2022.

\bibitem{DBLP:conf/cidr/ChengC07}
Tao Cheng and Kevin~Chen{-}Chuan Chang.
\newblock Entity search engine: Towards agile best-effort information
  integration over the web.
\newblock In {\em {CIDR}}, pages 108--113. www.cidrdb.org, 2007.

\bibitem{DBLP:conf/naacl/DevlinCLT19}
Jacob Devlin, Ming{-}Wei Chang, Kenton Lee, and Kristina Toutanova.
\newblock {BERT:} pre-training of deep bidirectional transformers for language
  understanding.
\newblock In {\em {NAACL-HLT} {(1)}}, pages 4171--4186. Association for
  Computational Linguistics, 2019.

\bibitem{DBLP:conf/iclr/DosovitskiyB0WZ21}
Alexey Dosovitskiy, Lucas Beyer, Alexander Kolesnikov, Dirk Weissenborn,
  Xiaohua Zhai, Thomas Unterthiner, Mostafa Dehghani, Matthias Minderer, Georg
  Heigold, Sylvain Gelly, Jakob Uszkoreit, and Neil Houlsby.
\newblock An image is worth 16x16 words: Transformers for image recognition at
  scale.
\newblock In {\em {ICLR}}. OpenReview.net, 2021.

\bibitem{DBLP:conf/cvpr/DouXGWWWZZYP0022}
Zi{-}Yi Dou, Yichong Xu, Zhe Gan, Jianfeng Wang, Shuohang Wang, Lijuan Wang,
  Chenguang Zhu, Pengchuan Zhang, Lu~Yuan, Nanyun Peng, Zicheng Liu, and
  Michael Zeng.
\newblock An empirical study of training end-to-end vision-and-language
  transformers.
\newblock In {\em {CVPR}}, pages 18145--18155. {IEEE}, 2022.

\bibitem{DBLP:conf/conll/EshelCRMYL17}
Yotam Eshel, Noam Cohen, Kira Radinsky, Shaul Markovitch, Ikuya Yamada, and
  Omer Levy.
\newblock Named entity disambiguation for noisy text.
\newblock In {\em CoNLL}, pages 58--68. Association for Computational
  Linguistics, 2017.

\bibitem{DBLP:conf/www/FangC0ZZL19}
Zheng Fang, Yanan Cao, Qian Li, Dongjie Zhang, Zhenyu Zhang, and Yanbing Liu.
\newblock Joint entity linking with deep reinforcement learning.
\newblock In {\em {WWW}}, pages 438--447. {ACM}, 2019.

\bibitem{DBLP:conf/naacl/Francis-LandauD16}
Matthew Francis{-}Landau, Greg Durrett, and Dan Klein.
\newblock Capturing semantic similarity for entity linking with convolutional
  neural networks.
\newblock In {\em {HLT-NAACL}}, pages 1256--1261. The Association for
  Computational Linguistics, 2016.

\bibitem{DBLP:conf/mm/GanLWWHH21}
Jingru Gan, Jinchang Luo, Haiwei Wang, Shuhui Wang, Wei He, and Qingming Huang.
\newblock Multimodal entity linking: {A} new dataset and {A} baseline.
\newblock In {\em {ACM} Multimedia}, pages 993--1001. {ACM}, 2021.

\bibitem{DBLP:conf/sigir/GerritseHV22}
Emma~J. Gerritse, Faegheh Hasibi, and Arjen~P. de~Vries.
\newblock Entity-aware transformers for entity search.
\newblock In {\em {SIGIR}}, pages 1455--1465. {ACM}, 2022.

\bibitem{DBLP:conf/emnlp/GuptaSR17}
Nitish Gupta, Sameer Singh, and Dan Roth.
\newblock Entity linking via joint encoding of types, descriptions, and
  context.
\newblock In {\em {EMNLP}}, pages 2681--2690. Association for Computational
  Linguistics, 2017.

\bibitem{DBLP:conf/icml/KimSK21}
Wonjae Kim, Bokyung Son, and Ildoo Kim.
\newblock Vilt: Vision-and-language transformer without convolution or region
  supervision.
\newblock In {\em {ICML}}, volume 139 of {\em Proceedings of Machine Learning
  Research}, pages 5583--5594. {PMLR}, 2021.

\bibitem{DBLP:conf/acl/TitovL18a}
Phong Le and Ivan Titov.
\newblock Improving entity linking by modeling latent relations between
  mentions.
\newblock In {\em {ACL} {(1)}}, pages 1595--1604. Association for Computational
  Linguistics, 2018.

\bibitem{DBLP:conf/acl/LewisLGGMLSZ20}
Mike Lewis, Yinhan Liu, Naman Goyal, Marjan Ghazvininejad, Abdelrahman Mohamed,
  Omer Levy, Veselin Stoyanov, and Luke Zettlemoyer.
\newblock {BART:} denoising sequence-to-sequence pre-training for natural
  language generation, translation, and comprehension.
\newblock In {\em {ACL}}, pages 7871--7880. Association for Computational
  Linguistics, 2020.

\bibitem{DBLP:conf/nips/LiSGJXH21}
Junnan Li, Ramprasaath~R. Selvaraju, Akhilesh Gotmare, Shafiq~R. Joty, Caiming
  Xiong, and Steven~Chu{-}Hong Hoi.
\newblock Align before fuse: Vision and language representation learning with
  momentum distillation.
\newblock In {\em NeurIPS}, pages 9694--9705, 2021.

\bibitem{DBLP:conf/mm/LiaoM0HC18}
Lizi Liao, Yunshan Ma, Xiangnan He, Richang Hong, and Tat{-}Seng Chua.
\newblock Knowledge-aware multimodal dialogue systems.
\newblock In {\em {ACM} Multimedia}, pages 801--809. {ACM}, 2018.

\bibitem{DBLP:journals/corr/abs-1907-11692}
Yinhan Liu, Myle Ott, Naman Goyal, Jingfei Du, Mandar Joshi, Danqi Chen, Omer
  Levy, Mike Lewis, Luke Zettlemoyer, and Veselin Stoyanov.
\newblock Roberta: {A} robustly optimized {BERT} pretraining approach.
\newblock {\em CoRR}, abs/1907.11692, 2019.

\bibitem{DBLP:conf/emnlp/LongprePCRD021}
Shayne Longpre, Kartik Perisetla, Anthony Chen, Nikhil Ramesh, Chris DuBois,
  and Sameer Singh.
\newblock Entity-based knowledge conflicts in question answering.
\newblock In {\em {EMNLP} {(1)}}, pages 7052--7063. Association for
  Computational Linguistics, 2021.

\bibitem{DBLP:conf/iclr/LoshchilovH19}
Ilya Loshchilov and Frank Hutter.
\newblock Decoupled weight decay regularization.
\newblock In {\em {ICLR} (Poster)}. OpenReview.net, 2019.

\bibitem{DBLP:conf/acl/CarvalhoMN18}
Seungwhan Moon, Leonardo Neves, and Vitor Carvalho.
\newblock Multimodal named entity disambiguation for noisy social media posts.
\newblock In {\em {ACL} {(1)}}, pages 2000--2008. Association for Computational
  Linguistics, 2018.

\bibitem{DBLP:conf/nips/PaszkeGMLBCKLGA19}
Adam Paszke, Sam Gross, Francisco Massa, Adam Lerer, James Bradbury, Gregory
  Chanan, Trevor Killeen, Zeming Lin, Natalia Gimelshein, Luca Antiga, Alban
  Desmaison, Andreas K{\"{o}}pf, Edward~Z. Yang, Zachary DeVito, Martin Raison,
  Alykhan Tejani, Sasank Chilamkurthy, Benoit Steiner, Lu~Fang, Junjie Bai, and
  Soumith Chintala.
\newblock Pytorch: An imperative style, high-performance deep learning library.
\newblock In {\em NeurIPS}, pages 8024--8035, 2019.

\bibitem{DBLP:conf/emnlp/PetersNLSJSS19}
Matthew~E. Peters, Mark Neumann, Robert L.~Logan IV, Roy Schwartz, Vidur Joshi,
  Sameer Singh, and Noah~A. Smith.
\newblock Knowledge enhanced contextual word representations.
\newblock In {\em {EMNLP/IJCNLP} {(1)}}, pages 43--54. Association for
  Computational Linguistics, 2019.

\bibitem{DBLP:conf/icml/RadfordKHRGASAM21}
Alec Radford, Jong~Wook Kim, Chris Hallacy, Aditya Ramesh, Gabriel Goh,
  Sandhini Agarwal, Girish Sastry, Amanda Askell, Pamela Mishkin, Jack Clark,
  Gretchen Krueger, and Ilya Sutskever.
\newblock Learning transferable visual models from natural language
  supervision.
\newblock In {\em {ICML}}, volume 139 of {\em Proceedings of Machine Learning
  Research}, pages 8748--8763. {PMLR}, 2021.

\bibitem{DBLP:conf/cvpr/SzegedyVISW16}
Christian Szegedy, Vincent Vanhoucke, Sergey Ioffe, Jonathon Shlens, and
  Zbigniew Wojna.
\newblock Rethinking the inception architecture for computer vision.
\newblock In {\em {CVPR}}, pages 2818--2826. {IEEE} Computer Society, 2016.

\bibitem{DBLP:conf/naacl/TsaiR16}
Chen{-}Tse Tsai and Dan Roth.
\newblock Cross-lingual wikification using multilingual embeddings.
\newblock In {\em {HLT-NAACL}}, pages 589--598. The Association for
  Computational Linguistics, 2016.

\bibitem{DBLP:conf/nips/VaswaniSPUJGKP17}
Ashish Vaswani, Noam Shazeer, Niki Parmar, Jakob Uszkoreit, Llion Jones,
  Aidan~N. Gomez, Lukasz Kaiser, and Illia Polosukhin.
\newblock Attention is all you need.
\newblock In {\em {NIPS}}, pages 5998--6008, 2017.

\bibitem{DBLP:journals/cacm/VrandecicK14}
Denny Vrandecic and Markus Kr{\"{o}}tzsch.
\newblock Wikidata: a free collaborative knowledgebase.
\newblock {\em Commun. {ACM}}, 57(10):78--85, 2014.

\bibitem{DBLP:journals/bdr/WangWQZ20}
Meng Wang, Haofen Wang, Guilin Qi, and Qiushuo Zheng.
\newblock Richpedia: {A} large-scale, comprehensive multi-modal knowledge
  graph.
\newblock {\em Big Data Res.}, 22:100159, 2020.

\bibitem{DBLP:conf/sigir/WangWC22}
Peng Wang, Jiangheng Wu, and Xiaohang Chen.
\newblock Multimodal entity linking with gated hierarchical fusion and
  contrastive training.
\newblock In {\em {SIGIR}}, pages 938--948. {ACM}, 2022.

\bibitem{DBLP:conf/acl/WangTGLWYCX22}
Xuwu Wang, Junfeng Tian, Min Gui, Zhixu Li, Rui Wang, Ming Yan, Lihan Chen, and
  Yanghua Xiao.
\newblock Wikidiverse: {A} multimodal entity linking dataset with diversified
  contextual topics and entity types.
\newblock In {\em {ACL} {(1)}}, pages 4785--4797. Association for Computational
  Linguistics, 2022.

\bibitem{DBLP:conf/www/WuZMGSH20}
Junshuang Wu, Richong Zhang, Yongyi Mao, Hongyu Guo, Masoumeh Soflaei, and
  Jinpeng Huai.
\newblock Dynamic graph convolutional networks for entity linking.
\newblock In {\em {WWW}}, pages 1149--1159. {ACM} / {IW3C2}, 2020.

\bibitem{DBLP:conf/emnlp/WuPJRZ20}
Ledell Wu, Fabio Petroni, Martin Josifoski, Sebastian Riedel, and Luke
  Zettlemoyer.
\newblock Scalable zero-shot entity linking with dense entity retrieval.
\newblock In {\em {EMNLP} {(1)}}, pages 6397--6407. Association for
  Computational Linguistics, 2020.

\bibitem{DBLP:conf/acl/XiongYCGW19}
Wenhan Xiong, Mo~Yu, Shiyu Chang, Xiaoxiao Guo, and William~Yang Wang.
\newblock Improving question answering over incomplete kbs with knowledge-aware
  reader.
\newblock In {\em {ACL} {(1)}}, pages 4258--4264. Association for Computational
  Linguistics, 2019.

\bibitem{DBLP:conf/conll/YamadaS0T16}
Ikuya Yamada, Hiroyuki Shindo, Hideaki Takeda, and Yoshiyasu Takefuji.
\newblock Joint learning of the embedding of words and entities for named
  entity disambiguation.
\newblock In {\em CoNLL}, pages 250--259. {ACL}, 2016.

\bibitem{DBLP:conf/emnlp/YangGLTZWCHR19}
Xiyuan Yang, Xiaotao Gu, Sheng Lin, Siliang Tang, Yueting Zhuang, Fei Wu,
  Zhigang Chen, Guoping Hu, and Xiang Ren.
\newblock Learning dynamic context augmentation for global entity linking.
\newblock In {\em {EMNLP/IJCNLP} {(1)}}, pages 271--281. Association for
  Computational Linguistics, 2019.

\bibitem{DBLP:conf/dasfaa/ZhangLY21}
Li~Zhang, Zhixu Li, and Qiang Yang.
\newblock Attention-based multimodal entity linking with high-quality images.
\newblock In {\em {DASFAA} {(2)}}, volume 12682 of {\em Lecture Notes in
  Computer Science}, pages 533--548. Springer, 2021.

\bibitem{DBLP:journals/dint/ZhengWWQ22}
Qiushuo Zheng, Hao Wen, Meng Wang, and Guilin Qi.
\newblock Visual entity linking via multi-modal learning.
\newblock {\em Data Intell.}, 4(1):1--19, 2022.

\end{thebibliography}

\clearpage

\appendix
\section{Appendix}

\begin{figure*}[t]
    \centering
    \includegraphics[width=0.95\textwidth]{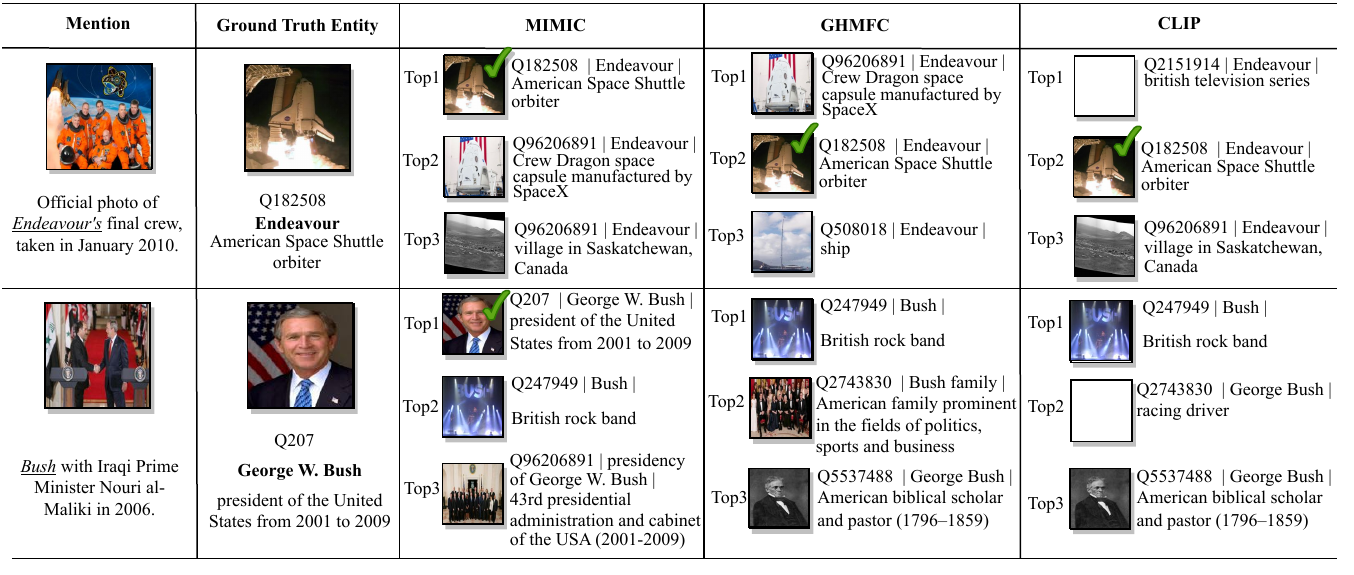}
    \caption{Case study for MEL. Each row is a case, which contains mention, ground truth entity, and top three retrieved entities of three methods, i.e., MIMIC (ours), GHMFC~\cite{DBLP:conf/sigir/WangWC22}, CLIP~\cite{DBLP:conf/icml/RadfordKHRGASAM21}. The \textit{italic} and \uline{underlined} words in mention are mention words. Each retrieved entity is described with three parts, Wikidata QID, entity name, a short description, and three parts are separated by "|". A blank square means that the corresponding entity has no image. The symbol "\color{teal}$\checkmark$ \color{black}" marks the correct entity.} 
    \label{fig:case_study}
\end{figure*}

\subsection{Details of datasets}
\label{sec:details_of_datasets}
\begin{figure}[t]
\centering
\includegraphics[width=0.475\textwidth]{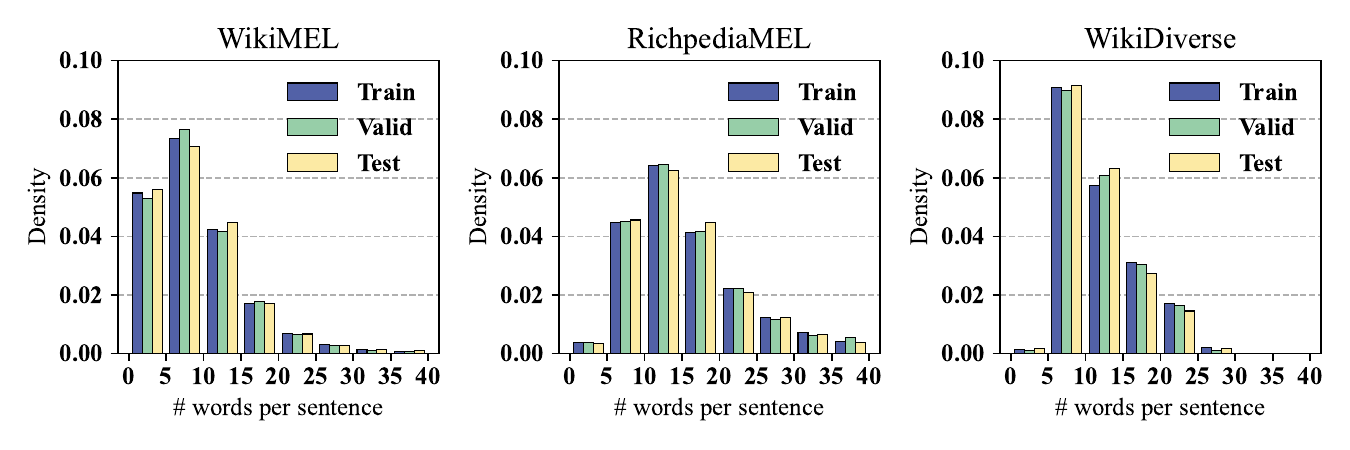}
\caption{Distribution of sentence length for three datasets.}
\label{fig:length}
\end{figure}

Table~\ref{tab:statistics} shows basic statistics of the three dataset.
Figure~\ref{fig:length} summarizes distribution of sentence length for the three
datasets, which indicates the balance among different splits.

\begin{table}[!t]
\centering
\caption{Statistics of three datasets. "Ment." and "sent." denote mention(s) and sentence(s) respectively.}
\begin{tabular}{c|ccc}
\hline
Statistic               & WikiMEL & RichpediaMEL & WikiDiverse \\ \hline
\# sentences            & 22,070   & 17,724        & 7,405        \\
\# mentions             & 25,846   & 17,805        & 15,093       \\
\# img. of  ment.       & 22,136   & 15,853        & 6,697        \\
\# ment. in train       & 18,092   & 12,463        & 11,351       \\
\# ment. in valid       & 2,585    & 1,780         & 1,664        \\
\# ment. in test        & 5,169    & 3,562         & 2,078        \\
\# entities of KB       & 109,976  & 160,935       & 132,460      \\
\# entities with img.   & 67,195   & 86,769        & 67,309       \\ \hline
\end{tabular}
\label{tab:statistics}
\end{table}

\subsection{Descriptions of Baselines}
\label{sec:desc_of_baselines}
We compared our proposed MIMIC with three groups of baselines. The first group of baselines is text-based methods.
\begin{itemize}[leftmargin=*]
    \item \textbf{BLINK}~\cite{DBLP:conf/emnlp/WuPJRZ20} is a two-stage zero-shot EL method and employs BERT as the backbone. It first retrieves entities with a bi-encoder and then re-ranks these candidate entities with a cross-encoder.
    \item \textbf{BERT}~\cite{DBLP:conf/naacl/DevlinCLT19} consists of a stack of Transformer encoders and is pre-trained on a large amount of corpus. BERT has shown the ability to solve many natural language understanding tasks. 
    \item \textbf{RoBERTa}~\cite{DBLP:journals/corr/abs-1907-11692} further improves BERT by removing the next sentence prediction objective and using a dynamic mask language model.
\end{itemize} 

The second group of baselines contains MEL method.
\begin{itemize}[leftmargin=*]
    \item \textbf{DZMNED}~\cite{DBLP:conf/acl/CarvalhoMN18} is the first method for MEL, which utilizes additional attention mechanism to fuse visual features, word-level textual features and char-level features.
    \item \textbf{JMEL}~\cite{DBLP:conf/ecir/AdjaliBFBG20} extracts both unigram and bigram embeddings as textual features. Different features are fused by concatenation and a fully connected layer. We replace the textual encoder with a pre-trained BERT for a fair comparison.
    \item \textbf{VELML}~\cite{DBLP:journals/dint/ZhengWWQ22} utilizes VGG-16 network to obtain object-level visual features. We use pre-trained BERT to replace the original GRU textual encoder. The two modalities are fused with additional attention mechanism.
    \item \textbf{GHMFC}~\cite{DBLP:conf/sigir/WangWC22} proposes hierarchical cross-attention to capture the underlying fine-grained correlation among textual and visual features and uses contrastive learning for optimization.
\end{itemize}

The third group of baselines includes Vision-and-Language Pre-training models.
\begin{itemize}[leftmargin=*]
    \item \textbf{CLIP}~\cite{DBLP:conf/icml/RadfordKHRGASAM21} employs two Transformer-based encoders to attain visual and textual representation, which pre-trains on massive noisy web data with contrastive loss.
    \item \textbf{ViLT}~\cite{DBLP:conf/icml/KimSK21} proposes to use shallow textual and visual embeddings, and concentrates on deep modality interaction via a stack of Transformer layers. 
    \item \textbf{ALBEF}~\cite{DBLP:conf/nips/LiSGJXH21} first aligns visual and textual features with image-text contrastive loss and then fuses them with a multimodal Transformer encoder. Momentum distillation is further applied to improve learning from noisy data.
    \item \textbf{METER}~\cite{DBLP:conf/cvpr/DouXGWWWZZYP0022} utilizes the co-attention schema to exploit the semantic relation of different modalities, where each layer consists of a self-attention module, cross-attention module and a feed-forward network.
\end{itemize}

\subsection{Evaluation Metrics}
\label{sec:eval_metric}
We first calculate the similarity scores between a mention and all entities of the KB, then the similarity scores are sorted in descending order to calculate \textbf{H@k}, \textbf{MRR} and \textbf{MR}, which are defined as:
\begin{gather}
    H@k = \frac{1}{N}\sum_{i}^{N}I(rank(i)<k), \\
    MRR = \frac{1}{N}\sum_{i}^{N}\frac{1}{rank(i)}, \\
    MR = \frac{1}{N}\sum_{i}^{N}rank(i),
\end{gather}
where $N$ is the number of total samples, $rank(i)$ means the rank of the i-th ground truth entity in the rank list of KB entities, $I(\cdot)$ stands for indicator function which is 1 if the subsequent condition is satisfied otherwise 0.

\subsection{Case Study}
For a more illustrative demonstration of the proposed MIMIC, we provided two cases and compared MIMIC with two strong competitors, i.e., GHMFC~\cite{DBLP:conf/sigir/WangWC22} and CLIP~\cite{DBLP:conf/icml/RadfordKHRGASAM21}, which is shown in Figure~\ref{fig:case_study}. In the first case, although three methods predict the correct entity in the top three retrieved entities, MIMIC distinguishes better between space shuttle and  space capsule by capturing the detailed information within the mention image. In the second case, two competitors retrieve rock band \textit{Bush} in the first place. MIMIC not only considers textual clues \textit{Bush} from the surface but also takes the visual scene of politics from the images into account, which helps to identify the correct entity.

\end{document}